\documentclass[conference]{IEEEtran}
\IEEEoverridecommandlockouts

\usepackage{cite}
\usepackage{amsmath,amssymb,amsfonts}
\usepackage{algorithmic}
\usepackage{graphicx}
\usepackage{textcomp}
\usepackage[table]{xcolor}
\usepackage{graphicx,verbatim}
\usepackage{amsmath}
\usepackage{amssymb}
\usepackage{booktabs}
\usepackage{amsfonts}
\usepackage{bm} 
\usepackage{graphicx}
\usepackage{multirow}
\usepackage{esvect}
\usepackage{seqsplit}
\def\BibTeX{{\rm B\kern-.05em{\sc i\kern-.025em b}\kern-.08em
    T\kern-.1667em\lower.7ex\hbox{E}\kern-.125emX}}
\begin{document}

\title{SpikeEEGNet: Spike-Driven Masked Autoencoder for Efficient EEG Modeling\\
}

\author{
Yongchang Gao$^{1}$,
Shihao Zou$^{2}$*,
Linxia Xiao$^{2}$,
Daiyan Li$^{3}$,
Qingqing Zheng$^{4}$,
Weixin Si$^{4}$, \IEEEmembership{Senior Member, IEEE}
\\
\IEEEauthorblockA{$^{1}$School of Computer Science and Engineering, University of Chinese Academy of Sciences, Beijing, China \\}
\IEEEauthorblockA{$^{2}$Shenzhen Institutes of Advanced Technology, Chinese Academy of Sciences, Shenzhen, China \\}
\IEEEauthorblockA{$^{3}$Southern University of Science and Technology, Shenzhen, China \\}
\IEEEauthorblockA{$^{4}$Shenzhen University of Advanced Technology, Shenzhen, China \\}
\IEEEauthorblockA{$*$ Corresponding author: sh.zou@siat.ac.cn \\}
}

\maketitle

\begin{abstract}
Electroencephalography (EEG) reflects the dynamics of underlying neuronal activity. However, most EEG decoding methods rely on artificial neural networks (ANNs), which use dense, continuous-valued computations that differ fundamentally from the discrete, spike-driven (0/1) communication of biological neural systems. Spiking neural networks (SNNs) provide a biologically inspired alternative by modeling spike timing explicitly, offering fine-grained spatiotemporal processing and potential advantages for transient EEG event prediction and brain-state classification. Nevertheless, their sparse spike representations can constrain feature expressiveness. To address this challenge, we propose SpikeEEGNet, a highly efficient SNN framework coupled with a self-supervised spike-driven masked autoencoder (SMAE) pretraining paradigm. By reconstructing randomly tube-masked EEG signals, the model learns robust spatiotemporal representations from unlabeled data and supports efficient downstream fine-tuning and personalized clinical adaptation. Experiments on five datasets, including three motor-imagery benchmarks and two early seizure-prediction cohorts--one of which is a private clinical cohort comprising 20 subjects--show that SpikeEEGNet consistently matches or surpasses state-of-the-art (SOTA) ANN-based methods. It achieves an ultra-low inference latency of $19.9\,\mu\text{s}$, less than 10\% of that of recent Transformer-based ANNs, while attaining SOTA accuracy of 76.8\% on the large-scale OpenBMI dataset. In the challenging early seizure prediction task on our private clinical cohort, SpikeEEGNet achieves the highest AUROC of 0.734 and improves sensitivity to 67.8\%, demonstrating its potential for efficient clinical EEG monitoring. 
\end{abstract}


\section{Introduction}
EEG non-invasively records brain activity through scalp electrodes, providing millisecond-level access to neural dynamics~\cite{lawhern2018eegnet}. Its rich spatiotemporal structure has made EEG a foundation of brain–computer interfaces (BCIs), enabling applications such as motor imagery classification, early seizure prediction, and cognitive workload assessment~\cite{song2022eeg,craik2019deep}. Since its discovery, EEG has been recognized as reflecting the collective, spike-driven activity of neuronal populations, where information is encoded through dynamic temporal interactions rather than static patterns~\cite{schirrmeister2017deep,altaheri2022physics}.

In recent years, ANNs have become the dominant paradigm for EEG decoding. Compact yet powerful architectures~\cite{lawhern2018eegnet,schirrmeister2017deep} achieve SOTA performance by learning hierarchical features from raw or minimally processed signals. However, ANNs rely on dense, continuous-valued activations and large-scale multiply–accumulate operations, which are fundamentally misaligned with spike-driven communication of neural activity~\cite{2019Towards}. This mismatch results in substantial computational redundancy and energy consumption, limiting their practicality for clinical applications where low latency and power efficiency are essential~\cite{kumar2022decoding}. Notably, the human brain achieves remarkable cognitive capability at roughly 20 watts~\cite{li2024brain,2025Neuromorphic}. This biological grounding suggests that decoding models should align with spike-based neural computation, motivating methods that are both accurate and efficient.

Emerging SNNs offer a neuro-inspired alternative by emulating spike-driven communication and enabling sparse, event-based computation. This paradigm naturally aligns with the temporal dynamics of EEG and promotes energy-efficient BCI systems. Unlike traditional ANNs, which may overlook fine-grained temporal patterns, SNNs explicitly encode precise spike timing, providing enhanced spatiotemporal modeling capability. Such advantages have recently enabled 100\% sensitivity in detecting transient EEG ``chirps''~\cite{davidhi2025event} and improved performance in brain state classification compared with conventional approaches~\cite{app14198911}.

However, SNNs remain underexplored in large-scale EEG decoding and often lag behind ANNs in performance. A central challenge is the intrinsic sparsity and discreteness of spike representations, which can cause information loss~\cite{neftci2019surrogate}. To bridge this gap, we propose SpikeEEGNet, an SNN-based framework that enhances representational capacity while maintaining energy efficiency. It combines a hierarchical spike-driven convolutional encoder with a self-supervised SMAE pretraining scheme. Through random tube masking to EEG's temporal structure, the model learns robust spatiotemporal representations from unlabeled data. The pre-trained encoder is subsequently fine-tuned for downstream EEG tasks or rapidly adapted for personalized diagnosis and treatment based on patient-specific EEG signals.

\begin{figure*}[t]
    \centering
    \includegraphics[width=\textwidth]{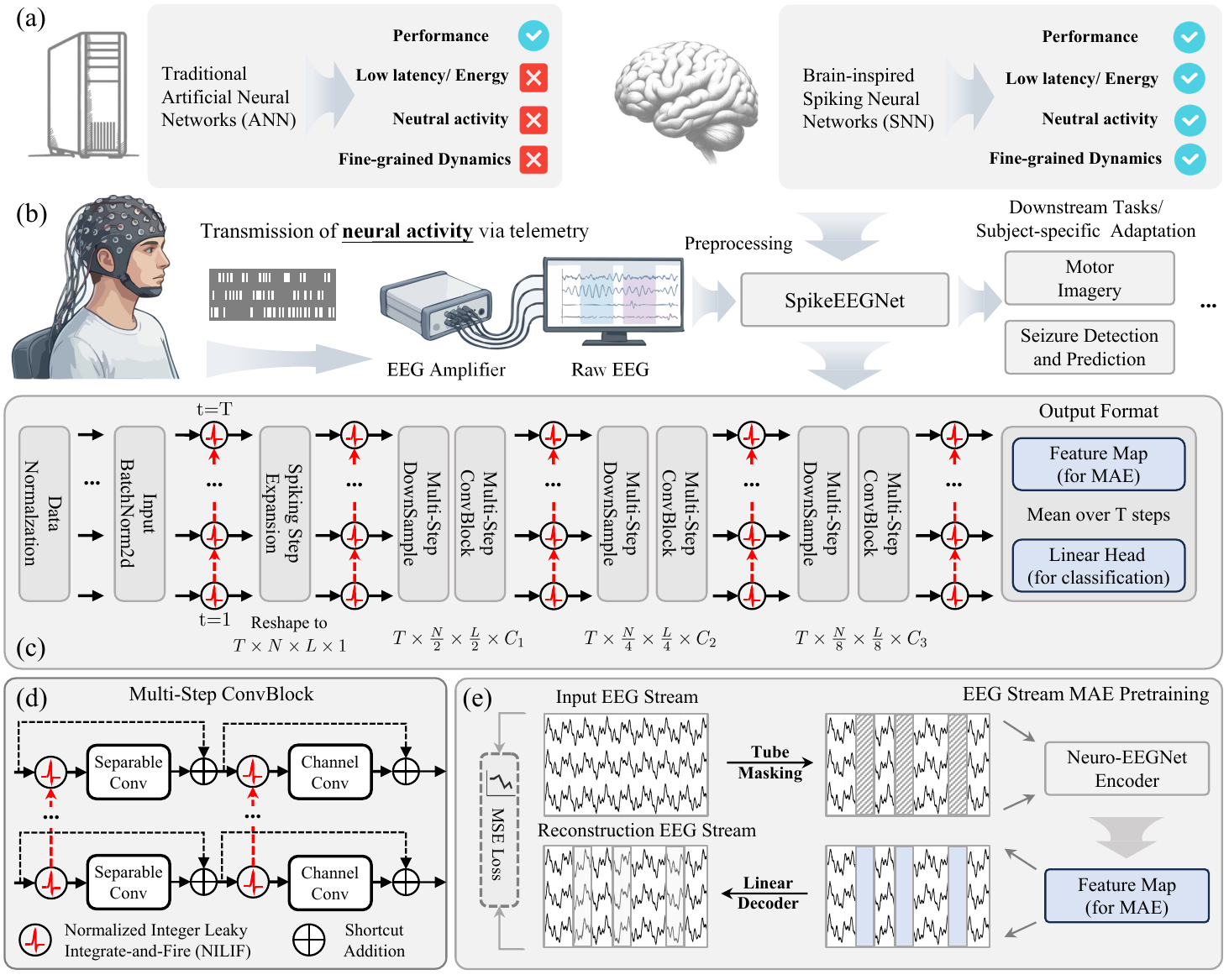}
    \caption{
    \textbf{SpikeEEGNet Overview.} \textbf{(a)} Conceptual comparison between conventional ANN-based EEG decoders and brain-inspired SNNs, highlighting the potential of spike-driven computation for low-latency, energy-efficient, and fine-grained neural-dynamics modeling. \textbf{(b)} Pipeline of EEG processing, from neural-signal acquisition and preprocessing to downstream decoding tasks and subject-specific adaptation. \textbf{(c)} Architecture of the SpikeEEGNet encoder, comprising data normalization, spiking-step expansion, hierarchical multi-step downsampling and ConvBlock stages, and temporal aggregation to produce representations for masked-autoencoder pretraining or downstream classification. \textbf{(d)} Design of the ConvBlock, which integrates spiking neurons, separable and channel-wise convolutions, and shortcut connections. \textbf{(e)} Self-supervised masked-autoencoder pretraining, in which tube-masked EEG signals are encoded and reconstructed by a lightweight linear decoder; the pretrained encoder is subsequently transferred to downstream tasks.
    }
    \label{fig:whole_method}
\end{figure*}

In summary, our contributions are threefold. 1) We introduce SpikeEEGNet, a highly efficient SNN-based framework for EEG decoding, effectively bridging biological plausibility with deep learning expressiveness while maintaining low computational cost. 2) We propose a self-supervised SMAE for EEG decoding with SNNs, enabling robust spatiotemporal representation learning from large-scale unlabeled data under sparse spike dynamics, while supporting efficient downstream fine-tuning and rapid personalized clinical adaptation. 3) Across five benchmarks, our method achieves competitive or superior performance while maintaining high computational efficiency. It attains SOTA accuracy of 76.8\% on the large-scale OpenBMI dataset with an ultra-low inference latency of $19.9\,\mu\text{s}$, less than 10\% of that of comparable ANN-based models. On the private clinical cohort, the model shows robust performance for preictal seizure prediction, achieving the highest overall AUROC of 0.734 and a sensitivity of 67.8\%. These results highlight its potential for efficient and clinically practical EEG monitoring.

\section{Related Works}

\noindent\textbf{EEG Modeling.} Deep learning has substantially advanced EEG decoding by learning spatio-temporal representations directly from raw or minimally processed signals. CNN-based architectures, such as EEGNet~\cite{lawhern2018eegnet} and DeepConvNet~\cite{schirrmeister2017deep}, provide effective and compact baselines, whereas attention-based models, including ATCNet~\cite{altaheri2022physics} and EEG-Conformer~\cite{song2022eeg}, further capture long-range temporal dependencies and cross-channel interactions. However, these supervised approaches depend heavily on high-quality annotations, whose acquisition is costly and time-consuming, particularly for medical EEG. To reduce this reliance on labeled data, self-supervised learning (SSL) methods, including contrastive learning and masked autoencoding, have been increasingly explored for EEG representation learning~\cite{kostas2021bendr,chien2022maeeg,he2022masked}. In particular, masked signal reconstruction enables models to learn robust representations from unlabeled EEG trials by inferring missing temporal content from the observed context. Nevertheless, existing EEG SSL frameworks are predominantly developed for ANN-based architectures, which rely on dense, continuous-valued computations and therefore incur substantial computational cost while remaining less aligned with spike-based neural communication~\cite{craik2019deep}. How to perform effective masked representation learning directly in the spiking domain remains underexplored. Addressing this gap is important for improving the representation capacity of SNN-based EEG decoders while preserving their sparse, event-driven efficiency.

\noindent\textbf{Spiking Neural Networks.} SNNs offer a biologically inspired and energy-efficient alternative to ANNs through binary spike communication and sparse computation~\cite{2025Neuromorphic,li2024brain}. Early approaches relied on ANN-to-SNN conversion, often incurring severe accuracy loss. With surrogate gradient training~\cite{neftci2019surrogate}, deep SNNs have achieved competitive performance in a wide range of vision tasks~\cite{yao2024spikev2,luo2024integer,zou2025spikevideoformer}. More recently, SNNs have emerged as a promising approach for EEG signal processing, as their precise spike timing naturally captures the dynamic, spatiotemporal nature of brain activity. For instance, in EEG-based emotion recognition, Kasabov \textit{et al.}~\cite{kasabov2014neucube} introduced the NeuCube architecture, which leverages unsupervised spike-timing-dependent plasticity (STDP) to model spatiotemporal patterns. Similarly, Gong \textit{et al.}~\cite{gong2023spiking} proposed SGLNet, integrating adaptive graph convolutions and LSTM units to effectively capture both spatial and temporal EEG dependencies. While these studies show the promise of SNNs for specialized EEG tasks, their application to high-dimensional, noisy signals across diverse paradigms remains challenging, partly due to sparse spike representations.

\section{Method}

\subsection{SNN Preliminary.} 

SNNs typically consist of stacked layers of spiking neurons, such as the Leaky Integrate-and-Fire (LIF) neurons. For the $i$-th layer with $N^{(i)}$ neurons, each neuron maintains a membrane potential $U^{(i)}_{[t]} \in \mathbb{R}^{N^{(i)}}$ that decays over time with a leakage factor $\beta$. At each spiking step $t$, the membrane potential is updated by charging with the weighted input spikes $W^{(i)} S^{(i-1)}_{[t]}$ received from the previous layer over the $T$ spiking steps. When the membrane potential exceeds a predefined threshold $u_{\text{th}}$, the neuron triggers a spike $S^{(i)}_{[t]}$ through the element-wise Heaviside step function $\Theta$ and undergoes a soft reset, subtracting $u_{\text{th}}$ from its potential~\cite{zhou2022spikformer}. The feedforward dynamics can be formulated as follows:
\begin{equation}
    \label{eq:feedforward}
    \begin{split}
        \tilde{U}_{[t]}^{(i)} &= \beta U^{(i)}_{[t-1]}+W^{(i)}S^{(i-1)}_{[t]}, \\
        S^{(i)}_{[t]} &= \Theta(\tilde{U}_{[t]}^{(i)}-u_{\rm th}), \\
        U^{(i)}_{[t]} &= \tilde{U}_{[t]}^{(i)}-u_{\rm th}S^{(i)}_{[t]}.
    \end{split}
\end{equation}
The computational efficiency of SNNs arises primarily from two properties. 1) \textit{Addition-based feedforward:} because the spike vector $S^{(i-1)}_{[t]}$ is binary, the charging operation $W^{(i)}S^{(i-1)}_{[t]}$ can be implemented through accumulation rather than dense, continuous-valued multiplication. 2) \textit{Sparse activation:} most entries in $S^{(i-1)}_{[t]}$ are inactive (approximately 80\% in our experiments), allowing the corresponding operations in $W^{(i)}S^{(i-1)}_{[t]}$ to be skipped.

\subsection{SNN Encoder.} 

Let $X^{\mathrm{in}} \in \mathbb{R}^{1 \times N \times L}$ denote an EEG sample, where $N$ is the number of electrodes and $L$ is the temporal length. We first reduce distributional differences in the raw recordings through Euclidean alignment and then apply Z-score normalization~\cite{he2020transfer}, yielding signals with zero mean and unit variance across channels and trials. Following direct encoding~\cite{yao2024spikev2}, the normalized EEG is repeated over $T$ spiking steps to construct the spatiotemporal input tensor $I \in \mathbb{R}^{T \times 1 \times N \times L}$ for subsequent spike-based processing.

To model EEG patterns across electrodes and time, the encoder first transforms the input into spiking feature maps using a patch-embedding layer, followed by three hierarchical stages of spike-driven ConvBlocks. Downsampling is applied between successive stages to halve both electrode and temporal resolutions while expanding the channel dimension. The output of stage $i$ is a spiking feature map of size $T \times C_i \times N_i \times L_i$, where $i \in \{1, 2, 3\}$. The spatial dimension $N_i$ and temporal dimension $L_i$ are progressively halved via stride-2 convolutions, bounded by a minimum size of 1, and $C_i$ denotes the stage-specific channel dimension. This progressive design aggregates local spatio-temporal information into increasingly abstract EEG representations. The membrane potentials from the final stage are then averaged over the $T$ simulation steps, producing a continuous latent representation $U_{\mathrm{latent}} \in \mathbb{R}^{C_3 \times N_3 \times L_3}$ for signal reconstruction and downstream classification. We do not incorporate spike-driven Transformer modules~\cite{yao2024spikev2,zou2025spikevideoformer}, as our empirical results suggest that a purely convolutional design converges faster, better preserves fine-grained local EEG dynamics, and incurs lower computational cost.

To enhance the representational capacity of SNNs for continuous and non-stationary EEG signals, we adopt the Normalized Integer Leaky Integrate-and-Fire (NILIF) neuron~\cite{luo2024integer,lei2025spike2former}. Unlike conventional binary LIF neurons, NILIF generates bounded multi-bit integer spikes (\textit{e.g.}, $0$ to $Z_{\text{max}}=4$) through quantized activation, enabling richer information transmission at each spiking step. Although NILIF is optimized using standard surrogate-gradient training, its inference can be equivalently unfolded into $Z_{\text{max}}$ sub-steps of binary spikes. Thus, the charging operation in Eq.~(\ref{eq:feedforward}) remains addition-based at every sub-step, preserving spike-driven efficiency without introducing costly multiplications.

\subsection{Masked Pretraining.}

\noindent\textbf{Random Tube Masking.} During SMAE pretraining, we employ tube masking to preserve the temporal continuity of neural events while preventing trivial reconstruction from unmasked electrodes at the same time points. Given a preprocessed EEG trial $X \in \mathbb{R}^{N \times L}$, we first sample a binary temporal mask $m \in \{0,1\}^{L}$ that masks a random subset of temporal segments with a predefined ratio $\alpha$ (\textit{e.g.}, $0.6$). The temporal mask is then broadcast across all $N$ electrodes to obtain $M \in \{0,1\}^{N \times L}$, such that signals from every channel are simultaneously removed at each masked time step. This design forces the model to reconstruct missing activity from the temporal context of visible segments and cross-channel dependencies across time. The masked input is defined as $X_{\text{masked}} = X \odot (1-M)$ and is subsequently reshaped and fed into the SNN encoder to obtain the latent representation $U_{\text{latent}} \in \mathbb{R}^{C_3 \times N_3 \times L_3}$.

\noindent\textbf{Lightweight ANN Decoding (Discarded after pretraining).}
We employ a highly lightweight ANN decoder $\mathcal{D}_{\phi}$ to encourage the SNN encoder to learn high-level semantic features rather than merely memorizing low-level details. The decoder takes the continuous latent representation $U_{\text{latent}}$ as input and progressively reconstructs the original EEG signal. To strictly limit the model's capacity and prevent shortcut learning, the decoding process is intentionally streamlined into three steps with precise dimensional transformations:
\begin{enumerate}
    \item \textit{Spatial Aggregation:} We collapse the spatial (electrode) dimension of the latent feature maps via global average pooling. This operation reduces the feature shape from $C_3 \times N_3 \times L_3$ to $C_3 \times L_3$, yielding a compact channel descriptor for each abstract time step.
    \item \textit{Temporal Upsampling:} To precisely match the original EEG temporal resolution $L$, we apply nearest-neighbor interpolation along the temporal axis, expanding the intermediate representation to $C_3 \times L$.
    \item \textit{Linear Projection:} A simple linear layer (implemented via a $1\times1$ convolution) projects the $C_3$ expanded embedding back to the original electrode space $N$ ($N \ll C_3$), generating the final reconstructed signal $\hat{X} \in \mathbb{R}^{N \times L}$.
\end{enumerate}

\noindent\textbf{Reconstruction Loss.}
The reconstruction loss is computed using the Mean Squared Error (MSE), evaluated exclusively on the masked regions:
\begin{equation}
    \mathcal{L}_{\text{SMAE}} = \frac{\left\| (X - \hat{X}) \odot M \right\|_2^2}{\| M\|_1}.
\end{equation}
This specific formulation incentivizes the encoder to capture the global dependencies and long-range temporal correlations that are essential for interpreting complex, non-stationary EEG patterns.

\section{Experiments}

We evaluate SpikeEEGNet on three motor imagery (MI) benchmarks and two seizure prediction cohorts, followed by ablations of SMAE pretraining, encoder depth, and NILIF configuration.

\subsection{Motor Imagery Classification}
\noindent\textbf{Task Setup and Datasets.} We evaluate MI classification on three public benchmarks: \textbf{OpenBMI} (54 subjects, 2 classes, 62 channels), \textbf{BCI IV 2a} (9 subjects, 4 classes, 22 channels), and \textbf{BCI IV 2b} (9 subjects, 2 classes, 3 channels). OpenBMI uses 20 motor-area channels with 4-second epochs at 100~Hz ($L=400$). BCI IV 2a uses all 22 channels with 4-second epochs at 250~Hz ($L=1000$). BCI IV 2b uses C3, Cz, and C4 channels with 3.0--7.5~s post-cue segments sampled at 100~Hz ($L=450$).

\noindent\textbf{Evaluation Protocols.} Following prior work~\cite{wang2025dbconformer}, we use the subject-dependent \textit{Chronological Order} (CO) protocol, assigning each subject's first 80\% of trials to training and the remaining 20\% to testing. For subject-independent evaluation, \textit{Leave-One-Subject-Out} (LOSO) cross-validation reserves one subject for testing and uses all others for training. Classification \textit{accuracy} (Acc) is the primary metric. In addition, to evaluate deployment feasibility, we report parameters, GFLOPs, energy consumption, and inference latency.

\noindent\textbf{Efficiency Estimation.} We estimate energy consumption in \textit{milli-joules} (mJ) and inference latency in \textit{micro-seconds} ($\mu$s) following~\cite{yao2024spikev2,zou2025spikevideoformer}. The estimates are derived from the floating-point operations (FLOPs) of ANN and SNN inference. Under 45,nm CMOS technology~\cite{horowitz20141}, a 32-bit multiply-and-accumulate (MAC) operation in ANNs consumes $E_{\text{MAC}}=4.6,\text{pJ}$, whereas an accumulate (AC) operation in SNNs requires only $E_{\text{AC}}=0.9,\text{pJ}$. Since one MAC corresponds to two FLOPs, the energy consumption is estimated as
\begin{equation}
\text{Energy}_{\text{ANN}}\,(\mathrm{mJ}) = \frac{\text{FLOPs}}{2} \times E_{\text{MAC}},
\end{equation}
\begin{equation}
\text{Energy}_{\text{SNN}}\,(\mathrm{mJ}) = \text{FLOPs} \times E_{\text{AC}}.
\end{equation}
Furthermore, inference latency is estimated on the AMD Xilinx ZCU104 FPGA using the FireFly accelerator~\cite{li2023firefly}. This platform delivers a throughput of 691.2\,G MAC-FLOPs/s for dense ANNs and an ultra-high 5529.6\,G AC-FLOPs/s for sparse SNNs. The theoretical latency is computed as:
\begin{equation}
    \text{Latency}\,(\mu\text{s}) = \frac{\text{FLOPs}}{\text{Throughput (FLOPs/s)}} \times 10^{6}.
\end{equation}

\begin{table*}[htbp]
    \centering
    \caption{
    Accuracy results for motor imagery classification under the chronological-order (CO) / leave-one-subject-out (LOSO) protocols. The compared methods include ANN-based baselines and two SNN-based models. \textbf{Bold} and \underline{underlining} indicate the best and second-best results, respectively.
    }
    \label{tab:overall-comparison}
    \renewcommand{\arraystretch}{1.1}
    \resizebox{0.85\textwidth}{!}{
    \begin{tabular}{llccc cccc}
        \toprule
            Type & Methods & OpenBMI & BCI IV 2a & BCI IV 2b & Params & GFLOPs & Energy & Latency \\
        \midrule
            \multirow{6}{*}{ANN} & DeepConvNet~\cite{schirrmeister2017deep} & 66.8/73.2 & 59.6/73.3 & 62.4/62.6 & 0.32M & 0.06 & 0.13 & 43.4 \\
            & ATCNet~\cite{altaheri2022physics} & 69.5/68.6 & 76.0/73.5 & 71.0/65.6 & - & - & - & - \\
            & EEG-Conformer~\cite{song2022eeg} & 71.8/70.1 & 78.6/73.1 & 72.0/64.2 & 0.81M & 0.22 & 0.51 & 159.1 \\
            & EEGNet~\cite{lawhern2018eegnet} & 65.5/\underline{73.3} & 69.9/73.6 & 68.4/\underline{67.8} & \underline{0.12M} & \underline{0.05} & 0.11 & 36.2 \\
            & MSVTNet~\cite{liu2024msvtnet} & 70.2/72.7 & 81.3/74.6 & \underline{73.6}/67.4 & - & - & - & - \\
            & MSCFormer\cite{zhao2025multi} & \underline{75.6}/\underline{73.3} & \underline{81.8}/\underline{77.0} & 72.2/64.7 & 0.67M & 0.33 & 0.76 & 238.7 \\
            
            \midrule
            \multirow{3}{*}{SNN} & Spikeformer~\cite{zhou2022spikformer} & 69.8/72.6 & 79.8/76.4 & 72.2/66.6 & 0.45M & 0.13 & 0.11 & 23.6 \\
            & NeuCube~\cite{kasabov2014neucube} & 67.9/69.4 & 61.0/70.2 & 60.1/59.4 & \textbf{0.07M} & \textbf{0.03} & \textbf{0.02} & \textbf{4.6}\\
            & \textbf{SpikeEEGNet} & \textbf{76.8/73.8} & \textbf{88.9/78.9} & \textbf{73.7/68.5} & 0.29M & 0.11 & \underline{0.10} & \underline{19.9} \\
        \bottomrule
    \end{tabular}}
\end{table*}

\noindent\textbf{Implementation Details.} The proposed framework is implemented in PyTorch with SpikingJelly~\cite{fang2023spikingjelly} and trained on a single NVIDIA A100 GPU using a two-stage strategy. Specifically, the spiking encoder operates over $T=8$ spiking steps and utilizes a hierarchical architecture comprising three spike-driven convolutional blocks with channel capacities of $C_1=32$, $C_2=64$, and $C_3=128$, respectively. The NILIF configuration is set to $Z_{\text{max}}=2$. In Stage 1, the encoder is pre-trained for 150 epochs with the SMAE objective using the AdamW optimizer with a learning rate of $1\times 10^{-3}$ and a cosine annealing schedule. An adaptive tube masking ratio $\alpha$ is used for each dataset: 0.75 for BCI IV 2a, 0.60 for OpenBMI, and 0.50 for BCI IV 2b. The checkpoint with the lowest reconstruction loss is selected for downstream initialization. In Stage 2, the encoder is fine-tuned for up to 20 epochs using label-smoothed cross-entropy loss.

\noindent\textbf{Results.} Tab.~\ref{tab:overall-comparison} compares SpikeEEGNet with six ANN and two SNN baselines. SpikeEEGNet achieves the highest accuracy across the three MI datasets under both protocols, reaching 76.8\% on OpenBMI under CO. On BCI IV 2a, it achieves 88.9\% under CO and 78.9\% under LOSO, exceeding MSCFormer by 7.1 and 1.9 percentage points, respectively. Estimated energy consumption and latency are 0.10~mJ and 19.9~$\mu$s per inference, with approximately $8.0\times$ and $12.0\times$ lower latency than EEG-Conformer and MSCFormer. It also exceeds SpikeFormer by 9.1 percentage points on BCI IV 2a and 7.0 points on OpenBMI under CO. Under LOSO, SpikeEEGNet reaches 73.8\% on OpenBMI and 68.5\% on BCI IV 2b, compared with the strongest ANN results of 73.3\% and 67.8\%, respectively. Together with the BCI IV 2a results, these findings show that the accuracy advantage extends to unseen subjects, although its magnitude varies across datasets and evaluation protocols.

\begin{figure*}[htb]
  \centering
  \includegraphics[width=\linewidth]{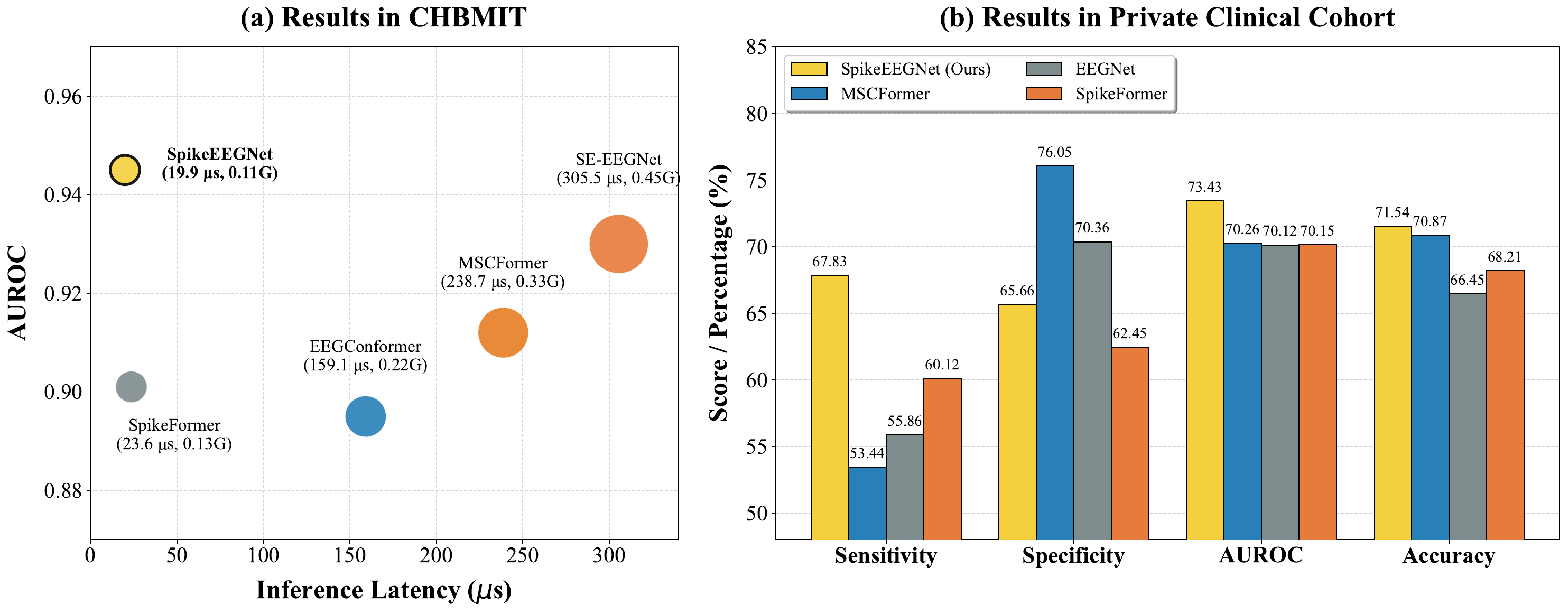}
  \caption{\textbf{Performance comparison on epileptic seizure prediction.} \textbf{(a)} Bubble chart illustrating the trade-off between AUROC and inference latency on the CHB-MIT dataset. The relative size of each bubble corresponds to the computational cost (FLOPs) of the model. \textbf{(b)} Comparison of clinical evaluation metrics (Sensitivity, Specificity, AUROC and Accuracy) across different models on the private clinical cohort. The grouped bar chart shows that ANN-based baselines tend to over-predict the majority class, whereas SpikeEEGNet achieves a more clinically viable balance across all metrics.}
  \label{fig:private dataset}
\end{figure*}

\subsection{Epileptic Seizure Prediction}
\noindent\textbf{Task Setup and Datasets.} We evaluate seizure prediction on \textbf{CHB-MIT}~\cite{CHBMIT}, which contains recordings from 23 pediatric patients, and a \textbf{private clinical cohort} of 20 patients with drug-resistant focal epilepsy. The private cohort comprises 60 seizure events and 56.29 hours of continuous EEG. All patients underwent surgical treatment with clinically validated seizure localization. Following prior work~\cite{qi2026epileptic,MAIWALD2004357}, positive samples are 4-second preictal windows drawn from 31 to 1 minutes before seizure onset, defining a 30-minute interval with a 1-minute warning horizon. Negative samples are interictal windows from seizure-free periods, excluding EEG within 4 hours before onset or 4 hours after seizure offset. For each seizure event, interictal windows are randomly sampled to approximately match the preictal sample count during supervised adaptation. Evaluation uses a non-overlapping 4-second stride for both preictal and interictal windows.

\noindent\textbf{Implementation Details.} For seizure prediction, each input sample comprises an EEG window with $N=18$ channels sampled at 256~Hz. Each 4-second window therefore contains $L=1024$ temporal samples. The seizure encoder uses $T=3$ spiking steps, NILIF neurons with $Z_{\mathrm{max}}=2$, and block-wise channel dimensions of $C_1=20$, $C_2=40$, and $C_3=64$. Task-specific pretraining is performed for 150 epochs with a masking ratio of 0.6. We use the AdamW optimizer with a learning rate of $1\times10^{-3}$ and a weight decay of 0.05, a batch size of 256 and cosine learning-rate annealing.

\noindent\textbf{Evaluation Protocols.} Following Xu et al.~\cite{xu2022multichannel}, we use \textit{within-subject leave-one-seizure-out} evaluation. For each patient with multiple seizures, one event is held out for testing in each fold, and the remaining events are used for downstream training and adaptation. Results are aggregated across folds and patients. We report Sensitivity, Specificity, and AUROC, together with inference latency and computational cost.

\noindent\textbf{Results on CHB-MIT.} Fig.~\ref{fig:private dataset}(a) shows that SpikeEEGNet achieves competitive predictive performance relative to ANN~\cite{song2022eeg,zhao2025multi} and SNN~\cite{zhou2022spikformer} baselines with reduced computational overhead. It achieves an AUROC of $0.945 \pm 0.016$ and a Sensitivity of $92.6\% \pm 2.4\%$, requiring 0.11 GFLOPs and an estimated inference latency of 19.9~$\mu$s.

\noindent\textbf{Results on Private Clinical Cohort.} On this challenging cohort with severe class imbalance, SpikeEEGNet achieves the highest AUROC of $0.7343 \pm 0.0685$ and Accuracy of $71.54\%$ (Fig.~\ref{fig:private dataset}(b)). Its Sensitivity reaches $67.83\% \pm 14.90\%$, compared with $53.44\% \pm 15.59\%$ for MSCFormer~\cite{zhao2025multi} and $55.86\% \pm 8.15\%$ for EEGNet~\cite{lawhern2018eegnet}, while maintaining a Specificity of $65.66\% \pm 9.14\%$. These results indicate a more balanced sensitivity--specificity profile than baselines that favor interictal predictions.

\subsection{Ablation Study}

We ablate SMAE pretraining, encoder depth, and NILIF configuration on BCI IV 2b under the CO protocol.

\begin{table}[t]
    \caption{Ablation of SMAE pretraining mask ratio on the BCI IV 2b dataset under the chronological-order (CO) protocol.}
    \label{tab:ablation_mask}
    \centering
    \renewcommand{\arraystretch}{1.1} 
    \begin{tabular*}{\linewidth}{@{\extracolsep{\fill}}lcc@{}}
        \toprule
        \textbf{Configuration} & \textbf{Accuracy (\%)} & \textbf{$\Delta$} \\
        \midrule
        w/o pretraining & 69.22 & \textit{-4.52} \\
        Mask Ratio: 0.30 & 73.04 & \textit{-0.70} \\ 
        Mask Ratio: 0.50 (Ours) & \textbf{73.74} & \textbf{--} \\
        Mask Ratio: 0.75 & 72.06 & \textit{-1.68} \\
        \bottomrule
    \end{tabular*}
\end{table}

\begin{table}[t]
    \caption{Ablation of the number of ConvBlocks on the BCI IV 2b dataset under the chronological-order (CO) protocol.}
    \label{tab:ablation_conv}
    \centering
    \renewcommand{\arraystretch}{1.1} 
    \begin{tabular*}{\linewidth}{@{\extracolsep{\fill}}lccc@{}}
        \toprule
        \textbf{Configuration} & \textbf{Accuracy (\%)} & \textbf{$\Delta$} & \textbf{Latency} \\
        \midrule
        1 ConvBlock  & 70.95 & \textit{-2.79} & 12.55 \\
        2 ConvBlocks & 72.90 & \textit{-0.84} & 16.20 \\
        3 ConvBlocks (Ours) & \textbf{73.74} & \textbf{--} & \textbf{19.90} \\
        \bottomrule
    \end{tabular*}
\end{table}

\begin{table}[t]
    \caption{Ablation of NILIF configuration ($Z_{max}$) on the BCI IV 2b dataset under the chronological-order (CO) protocol.}
    \label{tab:ablation_zmax}
    \centering
    \renewcommand{\arraystretch}{1.1} 
    \begin{tabular*}{\linewidth}{@{\extracolsep{\fill}}lcccc@{}}
        \toprule
        \textbf{Configuration} & \textbf{Accuracy (\%)} & \textbf{$\Delta$} & \textbf{Energy} & \textbf{Latency} \\
        \midrule
        $Z_{max} = 1$ & 72.67 & \textit{-1.07} & 0.077 & 15.38 \\
        $Z_{max} = 2$ (Ours) & \textbf{73.74} & \textbf{--} & \textbf{0.100} & \textbf{19.90} \\
        $Z_{max} = 4$ & 74.77 & \textit{+1.03} & 0.125 & 25.17 \\
        \bottomrule
    \end{tabular*}
\end{table} 

\noindent\textbf{SMAE Pretraining and Mask Ratio.} Removing SMAE pretraining reduces accuracy from 73.74\% to 69.22\%, a drop of 4.52 percentage points (Tab.~\ref{tab:ablation_mask}). Among the tested masking ratios, 0.50 performs best, exceeding 0.30 and 0.75 by 0.70 and 1.68 percentage points, respectively. All three masking ratios outperform training without pretraining, supporting the usefulness of masked EEG reconstruction for initializing the spiking encoder. The intermediate ratio achieves the strongest downstream performance, suggesting that the amount of visible temporal context matters alongside the presence of a reconstruction objective. Thus, both pretraining and the masking configuration contribute to the observed gains.

\noindent\textbf{SNN Encoder ConvBlocks.} Three ConvBlocks achieve the highest accuracy of 73.74\%, exceeding one and two blocks by 2.79 and 0.84 percentage points, respectively (Tab.~\ref{tab:ablation_conv}). The corresponding latency increases from 12.55 to 16.20 and 19.90~$\mu$s, illustrating the accuracy--latency trade-off. The largest improvement occurs when increasing the encoder from one to two blocks, while the third block provides a smaller additional gain. This pattern indicates diminishing accuracy returns with depth and motivates considering inference latency together with accuracy when selecting the encoder configuration.

\noindent\textbf{NILIF Configuration $Z_{max}$.} Relative to $Z_{max}=2$, binary spikes ($Z_{max}=1$) reduce accuracy by 1.07 percentage points, whereas $Z_{max}=4$ improves it by 1.03 points (Tab.~\ref{tab:ablation_zmax}). The latter increases energy by 25.0\% to 0.125~mJ and latency by 26.5\% to 25.17~$\mu$s. We therefore retain $Z_{max}=2$ as the default, balancing 73.74\% accuracy with estimated energy of 0.100~mJ and latency of 19.90~$\mu$s.

\section{Conclusion}
We present SpikeEEGNet, a neuro-inspired and efficient framework for EEG decoding. Across multiple motor imagery and seizure prediction benchmarks, the proposed approach achieves consistent predictive performance while significantly reducing computational overhead and latency compared to ANN-based models. These results suggest that spike-driven representation learning offers a viable pathway toward scalable, low-power, and brain-aligned EEG decoding systems for next-generation wearable and clinical neurotechnology.

A current limitation of this field concerns the physical deployment of SNN models on dedicated neuromorphic hardware. As noted in~\cite{2025Neuromorphic,neuromorphic_commercial_success}, neuromorphic platforms are still in an early stage of development, with algorithmic innovations typically advancing ahead of hardware maturity. Nevertheless, because our architecture shares fundamental design principles with the model in~\cite{2024Spike}, we anticipate similar feasibility for hardware implementation. Exploring and validating the deployment of SpikeEEGNet on specialized edge chips remains a primary direction for our future work.

\section*{Acknowledge}
This work was supported in part by the National Natural Science Foundation of China (Nos. 62372441 and 62303442), the Guangdong Basic and Applied Basic Research Foundation (No. 2026B1515020093), the Shenzhen Science and Technology Program (Grant Nos. \seqsplit{RCYX20231211090127030} and \seqsplit{JCYJ20250604182948064}), and the Research Program of the Department of Education of Guangdong Province (Grant No. 2025ZDZX3027).

\bibliographystyle{IEEEtran} 
\bibliography{references}

\end{document}